\documentclass{article}

\PassOptionsToPackage{numbers, compress}{natbib}

\usepackage[dblblindworkshop, final]{neurips_2026}

\workshoptitle{Transitioning from Pre-training to Post-training}

\usepackage[utf8]{inputenc} %
\usepackage[T1]{fontenc}    %
\usepackage{hyperref}       %
\usepackage{url}            %
\usepackage{fvextra}        %
\usepackage{booktabs}       %
\usepackage{amsfonts}       %
\usepackage{nicefrac}       %
\usepackage{microtype}      %
\usepackage{xcolor}         %

\usepackage{CJKutf8}
\usepackage{tikz}
\usetikzlibrary{positioning, arrows.meta}
\newcommand{\zh}[1]{\begin{CJK}{UTF8}{gbsn}#1\end{CJK}}

\definecolor{blueDot}{RGB}{155,193,255}
\definecolor{whiteDot}{RGB}{255,255,255}
\definecolor{greenDot}{RGB}{118,210,124}
\definecolor{redDot}{RGB}{255,135,145}
\definecolor{grayDot}{RGB}{180,180,180}
\definecolor{darkBlue}{RGB}{35,83,160}
\definecolor{lineGray}{RGB}{210,210,210}

\DeclareRobustCommand{\captdot}[1]{%
  \tikz[baseline=-0.6ex]{\node[circle,fill=#1,draw=gray!60,inner sep=0pt,minimum size=6pt]{};}}%

\title{Training LLMs to Follow Reasoning Instructions\\with Minimal Edits}

\author{%
  Alexandra Yakovleva \\
  Aalto University \\
  \texttt{alexandra.yakovleva@aalto.fi} \\
  \And
  Luat Gia Khoi Nguyen \\
  System 2 AI \\
  \And
  Juho Kannala \\
  Aalto University \\
  University of Oulu \\
  \And
  Harri Valpola \\
  System 2 AI \\
  \And
  Alexander Ilin \\
  System 2 AI \\
}

\begin{document}

\maketitle

\begin{abstract}

Large language models increasingly rely on extended reasoning traces, yet the training signals used to shape this behavior remain coarse. Supervised fine-tuning on full trajectories forces the model to imitate an entire demonstration, overconstraining its output distribution, while trajectory-level reinforcement learning provides only weak credit assignment, rewarding or penalizing whole completions without identifying where they succeed or fail. We propose \emph{edit-based training}, a general framework for learning from a model's own trajectories via localized textual corrections. Given a sampled completion, an editor identifies the first point of failure under a specified criterion and proposes a minimal replacement span. The model is then supervised only on these edited spans. By design, this concentrates supervision where the model's reasoning first breaks down and aims to preserve its existing reasoning patterns elsewhere. The method applies both offline and online, with the latter continually adapting supervision to the model's evolving failure modes.
We instantiate this approach on reasoning instruction following with Qwen3-32B on the ReasonIF benchmark. Compared to full-trajectory imitation and trajectory-level reinforcement learning baselines, edit-based training substantially improves compliance with the reasoning instruction while largely preserving the model's skills on untargeted tasks. These results suggest that localized correction is an effective complement to these approaches for shaping how LLMs reason.
\end{abstract}

\section{Introduction}

Large language models (LLMs) are increasingly trained to solve problems through extended reasoning. In many recent systems, reasoning traces are not directly supervised step by step; instead, they emerge from optimization procedures that reward successful final behavior. This indirect supervision is partly a practical necessity. Specifying high-quality reasoning traces is difficult: good reasoning is often task-dependent, may admit many valid forms, and is hard to supervise at scale. Moreover, effective reasoning must remain compatible with the knowledge, linguistic patterns, and latent capabilities already present in the pretrained model.

At the same time, control over the reasoning process is increasingly important. Reasoning is not merely a means of producing an answer; it is also where a model interprets the task, forms a plan, applies constraints, checks assumptions, and determines which intermediate conclusions to trust. Better control over these steps could improve reliability, alignment, and robustness. Concrete needs include articulating intermediate steps the model would otherwise compress, blocking known shortcuts, requesting verification before the answer, or constraining style for downstream auditing. Reliable reasoning instruction following is a prerequisite for using such constraints to make models more controllable and transparent while reducing shortcuts, hallucinations, and reward hacking inside reasoning traces~\citep{kwon2025reasonif}. In practice, however, current reasoning-capable LLMs often fail to follow such instructions; they omit required steps, ignore constraints, or deviate from a prescribed strategy even when the requirements are stated explicitly~\cite{kwon2025reasonif, fu2025mathif}, and open reasoning models comply on fewer than 25\% of ReasonIF traces. This motivates the central goal of our work: to make language models more reliably follow instructions about their reasoning process, not only instructions about the final answer.

A second motivation arises from how reasoning models are commonly trained. Reinforcement-learning methods for reasoning, including Group Relative Policy Optimization (GRPO)-style approaches~\cite{shao2024deepseekmath}, typically assign rewards or advantages to entire sampled trajectories. A reasoning trace is therefore encouraged or discouraged as a whole, often based on its final outcome. Such trajectory-level feedback can be inefficient. An incorrect trajectory may contain mostly sound reasoning, with failure caused by a single local mistake, such as a missed constraint, an invalid inference, or an arithmetic error. Penalizing the entire trajectory risks discouraging useful reasoning behavior together with the specific step that should be corrected. Conversely, a trajectory that reaches the correct answer may still contain unnecessary, brittle, or logically flawed intermediate steps; rewarding the full trajectory can reinforce these undesirable patterns simply because the final answer is correct.

These observations suggest that improving reasoning requires more targeted interventions. Instead of treating a reasoning trace as an indivisible object, we seek to identify the critical point at which the model first goes wrong and provide supervision only at that point. Our approach begins by sampling a reasoning trace from the model. We then evaluate the trace to locate the first violation of the reasoning instruction. Once this point is identified, we employ an editing model, or editor, to revise only the local span that contains the error.

The edited span becomes the training target. Rather than training the model to imitate an entire corrected trajectory, we apply the edit loss only to the small number of tokens needed to repair the mistake. This design is based on the observation that small local corrections can redirect the model’s subsequent reasoning. When a trace is repaired with just enough correct text at the critical point, the corrected tokens can steer the model away from an erroneous path and toward a better continuation.

As a result, the method applies minimal supervision and is designed to preserve much of the model’s original reasoning behavior. It intervenes only where the model’s reasoning first breaks down, allowing the remaining trajectory to be generated according to the model’s existing capabilities and priors. In this sense, our approach lies between reinforcement learning and conventional SFT. Like reinforcement learning, it provides sparse guidance rather than full demonstration trajectories. Unlike trajectory-level reinforcement learning, however, the guidance is textual and localized: it specifies what should have been produced at the precise point where the reasoning fails. We argue that this form of localized correction offers a practical route toward more controllable training while limiting changes unrelated to the specified failure.

Our contributions are threefold. First, we introduce edit-based training, a framework that converts model-generated reasoning failures into localized textual supervision at the first point of violation. Second, we instantiate the framework in offline and online variants for reasoning instruction following, with editors that can be scripted verifiers, LLM editors, or hybrids of the two.
Third, we compare edit-based training against full-trajectory SFT and trajectory-level GRPO, and show that it improves reasoning instruction following while largely preserving the base model's skills on the untargeted tasks we measure.

\section{Edit-based training for reasoning instruction following}

\subsection{Problem setting}

We study the problem of training language models to better follow instructions about their reasoning process. Our focus is not primarily on improving final task accuracy. Instead, we consider settings in which the user specifies constraints on how reasoning should proceed, and the model must satisfy those constraints while generating its solution. Examples of such reasoning-process instructions include requirements to use a prescribed format, follow a specified sequence of steps, check particular assumptions, avoid certain intermediate shortcuts, explicitly apply given constraints, reason in a particular language or style, or separate planning, computation, and final answering. We treat final performance improvement as a natural downstream application, but the central objective of this work is reasoning control: making the model’s intermediate reasoning more compliant with the requested procedure.

A natural approach is to construct full reasoning traces that satisfy the desired constraints and train the model using supervised fine-tuning (SFT). One option is to prompt a stronger model to generate complete compliant reasoning traces. However, this can introduce a distribution mismatch. The stronger model may have different linguistic habits, may use reasoning patterns unavailable to the target model, may solve the problem using different priors or latent capabilities, or may produce demonstrations that are formally correct but unnatural for the target model to imitate. Another option is to ask a stronger model to rewrite full traces produced by the target model so that they satisfy the instructions. This reduces the mismatch, but it still imposes complete corrected trajectories on the target model.

Full-trajectory SFT has a more fundamental limitation: it treats one complete reasoning trace as the desired behavior, even though many valid continuations may exist. As a result, it can unnecessarily reduce the entropy of the model’s reasoning distribution. The model is trained not only to avoid the mistake, but also to imitate many incidental choices in the demonstration. This problem is especially severe when the training set is small. With limited demonstrations, full-trajectory SFT can overfit to narrow reasoning styles or solution paths. In contrast, our goal is to fix the model’s behavior while avoiding this entropy-collapse problem~\cite{wang2026curiosft, chen2026sedsft}.

Another natural approach is to use reinforcement learning, such as GRPO-style training, with rewards based on whether the generated reasoning satisfies the specified constraints. This approach avoids full demonstrations and can optimize directly for instruction satisfaction. However, trajectory-level rewards provide coarse feedback. If a long reasoning trace violates an instruction because of one local mistake, the reward penalizes the entire trajectory. The central difficulty is credit assignment: terminal rewards indicate whether a completion succeeded, but not where it first failed or what local change would have repaired it. Our method addresses this limitation by replacing trajectory-level feedback with localized textual corrections at the first point of failure.

\subsection{Edit-based training procedure}
\label{subsec:method}

\begin{figure}[t]
\centering
\resizebox{\textwidth}{!}{%
\begin{tikzpicture}[
  font=\sffamily\small,
    title/.style={align=left},
  badge/.style={circle, fill=darkBlue, text=white, minimum size=7mm},
  dot/.style={circle, draw=gray!60, minimum size=5.5mm, inner sep=0pt},
  label/.style={anchor=west, font=\sffamily\small, align=left},
  col/.style={xshift=#1}
]

\def\ytitle{-0.3}
\def\yprompt{-1.4}
\def\ystepone{-2.1}
\def\ysteptwo{-2.8}
\def\ystepk{-3.5}
\def\ydots{-4.2}
\def\ystepn{-4.9}
\def\yanswer{-5.6}

\newcommand{\heading}[3]{%
  \node[title, anchor=west] at ([xshift=-2.75mm]#1,\ytitle) {#3};
}

\newcommand{\myitem}[5]{%
  \node[dot, fill=#3] at (#1,#2) {};
  \node[label, text=#4] at (#1+0.35,#2) {#5};
}

\newcommand{\myvline}[3]{%
  \draw[lineGray, line width=1.1pt] (#1,#2) -- (#1,#3);
}

\def\xA{0}
\def\xB{3.2}
\def\xC{6.4}
\def\xD{8.0}
\def\xE{10.0}
\def\xF{12.8}

\myvline{\xA}{\yprompt}{\yanswer}
\foreach \y/\txt in {
  \ystepone/{Okay{,} ...},
  \ysteptwo/{First{,} ...},
  \ystepk/{Actually{,} ...},
  \ystepn/{So{,} ...},
  \yanswer/{The answer ...}
}{
  \node[dot, fill=grayDot] at (\xA,\y) {};
  \node[label] at (\xA+0.35,\y) {\txt};
}
\node[dot, fill=blueDot] at (\xA,\yprompt) {};
\node[dot, fill=redDot] at (\xA,\ystepone) {};
\node[label] at (\xA+0.35,\ydots) {$\cdots$};

\myvline{\xB}{\yprompt}{\yanswer}
\foreach \y/\txt in {
  \ystepone/{\zh{好的} ...},
  \ysteptwo/{\zh{首先} ...}
}{
  \node[dot, fill=whiteDot] at (\xB,\y) {};
  \node[label] at (\xB+0.35,\y) {\txt};
}
\node[dot, fill=blueDot] at (\xB,\yprompt) {};
\node[dot, fill=greenDot] at (\xB,\ystepone) {};
\node[dot, fill=grayDot] at (\xB,\ysteptwo) {};
\node[dot, fill=redDot] at (\xB,\ystepk) {};
\node[label] at (\xB+0.35,\ystepk) {Hmm{,} ...};
\node[dot, fill=grayDot] at (\xB,\ystepn) {};
\node[label] at (\xB+0.35,\ystepn) {Now{,} ...};
\node[dot, fill=grayDot] at (\xB,\yanswer) {};
\node[label] at (\xB+0.35,\yanswer) {The answer ...};
\node[label] at (\xB+0.35,\ydots) {$\cdots$};

\myvline{\xC}{\yprompt}{\yanswer}
\foreach \y/\txt in {
  \ystepone/{\zh{好的} ...},
  \ysteptwo/{\zh{首先} ...}
}{
  \node[dot, fill=whiteDot] at (\xC,\y) {};
  \node[label] at (\xC+0.35,\y) {\txt};
}
\node[dot, fill=blueDot] at (\xC,\yprompt) {};
\node[dot, fill=greenDot] at (\xC,\ystepone) {};
\node[dot, fill=whiteDot] at (\xC,\ysteptwo) {};
\node[dot, fill=greenDot] at (\xC,\ystepk) {};
\node[label] at (\xC+0.35,\ystepk) {\zh{嗯} ...};
\node[dot, fill=grayDot] at (\xC,\ystepn) {};
\node[label] at (\xC+0.35,\ystepn) {Right{,} ...};
\node[dot, fill=grayDot] at (\xC,\yanswer) {};
\node[label] at (\xC+0.35,\yanswer) {The answer ...};
\node[label] at (\xC+0.35,\ydots) {$\cdots$};

\node[label] at (\xD+0.35,\ydots) {$\cdots$};

\myvline{\xE}{\yprompt}{\yanswer}
\foreach \y/\txt in {
  \ystepone/{\zh{好的} ...},
  \ysteptwo/{\zh{首先} ...}
}{
  \node[dot, fill=whiteDot] at (\xE,\y) {};
  \node[label] at (\xE+0.35,\y) {\txt};
}
\node[dot, fill=blueDot] at (\xE,\yprompt) {};
\node[dot, fill=greenDot] at (\xE,\ystepone) {};
\node[dot, fill=grayDot] at (\xE,\ystepn) {};
\node[label] at (\xE+0.35,\ystepn) {\zh{我们}...};
\node[dot, fill=greenDot] at (\xE,\ystepk) {};
\node[label] at (\xE+0.35,\ystepk) {\zh{嗯} ...};
\node[dot, fill=redDot] at (\xE,\yanswer) {};
\node[label] at (\xE+0.35,\yanswer) {\zh{答案} ...};
\node[label] at (\xE+0.35,\ydots) {$\cdots$};

\myvline{\xF}{\yprompt}{\yanswer}
\foreach \y/\txt in {
  \ystepone/{\zh{好的} ...},
  \ysteptwo/{\zh{首先} ...}
}{
  \node[dot, fill=whiteDot] at (\xF,\y) {};
  \node[label] at (\xF+0.35,\y) {\txt};
}
\node[dot, fill=blueDot] at (\xF,\yprompt) {};
\node[dot, fill=greenDot] at (\xF,\ystepone) {};
\node[dot, fill=whiteDot] at (\xF,\ystepn) {};
\node[label] at (\xF+0.35,\ystepn) {\zh{我们}...};
\node[dot, fill=greenDot] at (\xF,\ystepk) {};
\node[label] at (\xF+0.35,\ystepk) {\zh{嗯} ...};
\node[dot, fill=greenDot] at (\xF,\yanswer) {};
\node[label] at (\xF+0.35,\yanswer) {The answer ...};
\node[label] at (\xF+0.35,\ydots) {$\cdots$};

\end{tikzpicture}%
}%
\caption{Edit-based training within one example. The instruction asks
the model to reason in Chinese; the final answer is to remain in
English. Each column shows the state of the trajectory after one
editing round. \captdot{blueDot}~The prompt and instruction are fixed
across rounds. \captdot{redDot}~A red circle marks the first remaining
instruction violation at the current round. \captdot{greenDot}~A green
circle is the editor's minimal replacement span and is used as a
supervised training target. \captdot{grayDot}~Gray circles are
model-generated tokens that are uncorrected in the current round.
\captdot{whiteDot}~White circles are tokens accepted as part of the
final trajectory but not used as edit targets. In the sampled rollout, the
model begins reasoning in English, so the first reasoning
step is flagged as the first violation. The editor makes
a minimal Chinese replacement, and generation resumes from the corrected
prefix. The editor applies replacements until the model
continues reasoning in Chinese. Once the model adopts the requested
mode, it may overshoot and emit the final answer in Chinese as well, which is itself a violation.
The last edit then restores an English answer. The trajectory is now
compliant.}
\label{fig:method}
\end{figure}
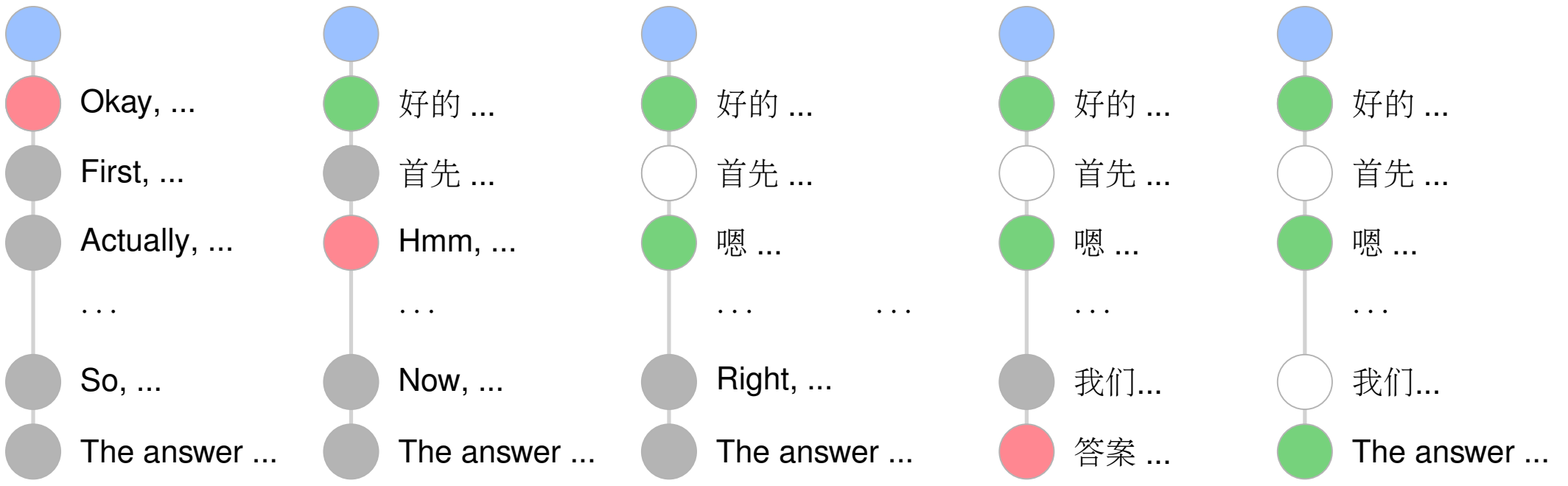

Given a prompt with an instruction about the desired reasoning process, we first sample a response from the current model (see~Figure~\ref{fig:method}). We then inspect the generated reasoning trace to find the first point at which it fails to follow the instruction. At that point, an editor proposes a minimal replacement span. The replacement is designed to be surgical: it changes only the local text needed to repair the failure. Such an edit may consist of only a few tokens, for example replacing an incorrect transition, inserting a missing constraint, correcting a computation, or changing a step that violates the required format. We intentionally avoid rewriting the entire reasoning trace.

After applying the edit, we continue sampling from the model conditioned on the corrected prefix. The corrected tokens may redirect the subsequent reasoning, allowing the model to recover and continue in a compliant way. We then evaluate the new continuation again, locate the first failure in the new continuation if one exists, and propose another local edit. This process can be repeated until the resulting trajectory satisfies the reasoning instruction or a stopping criterion is reached. The final output of the procedure is a single model-generated trajectory containing several short corrected spans. During training, the edit loss targets only the corrected spans. Thus, corrective supervision focuses on the counterfactual local actions that would have prevented the reasoning failure.

The editor is a module that receives a prompt, a generated trace, and the reasoning instruction. Its role is to identify the earliest span where the trace violates the instruction and to propose a replacement span. The editor may be implemented in different ways depending on the task: as a scripted verifier for formal or easily checkable constraints, as an LLM editor for more semantic reasoning instructions, or as a hybrid system combining rule-based checks with model-based editing. In all cases, the editor is instructed to make minimal edits rather than full rewrites. The proposed replacement is mapped to the model's tokens, and the first token that differs from the generated trace becomes the corrected span. If, after this token, the model's next generated token is the same as in the original trace, the span is extended token by token until the next generated token differs from the original trace. On average, a corrected span is shorter than two tokens.

Formally, let the model generate a reasoning trace conditioned on prompt $x$. Suppose the editor identifies a failure prefix $s_k$, where $s_k$ contains the prompt and the generated text up to the first failure point. The editor then provides a corrected replacement span $y_k^\star$. We train the model to assign higher likelihood to the corrected span at that failure state:
\begin{equation}
\ell_{\mathrm{edit}}(\pi)
=
\mathbb{E}_{s_k \sim D_{\mathrm{fail}}^\pi}
\left[
-\log \pi(y_k^\star \mid s_k)
\right].
\label{eq:editloss}
\end{equation}
Here, $D_{\mathrm{fail}}^\pi$ denotes the distribution of failure prefixes induced by the current policy $\pi$. The important distinction from standard supervised fine-tuning is that supervision is applied only at the edited span, not over the full trajectory. If multiple edits are found in one trajectory, the cross-entropy is averaged over the tokens of all edited spans. On the generated continuations between edits, we additionally apply a KL regularizer toward the base model (Appendix~\ref{app:training-configuration}).

We consider two ways of applying the edit-based training procedure. In the \emph{offline} setting, we first generate a fixed dataset of model completions. We run the editor on these completions to produce corrected spans, and then train the model on this fixed edit dataset. In the \emph{online} setting, we repeatedly generate completions using the current model. We then identify the current model’s failures, produce edits, and train on those edits. As the model improves, the distribution of failures changes. Online editing therefore focuses supervision on the mistakes that the current model actually makes. This makes the online variant near-on-policy: training data is drawn from the model’s evolving failure distribution.

\paragraph{Choice of loss function.}
We use cross-entropy on the edit span, as in (\ref{eq:editloss}). Treating the edit as a preference pair and optimizing the policy with Direct Preference Optimization (DPO)~\citep{rafailov2023direct} is a natural alternative (formulated in Appendix~\ref{app:dpo-loss}); we leave a systematic comparison to future work.

\paragraph{Relation to standard SFT.}
The key distinction from full-trajectory SFT is where supervision is concentrated. Our SFT baseline uses the same trajectories as Offline, treating both the model's own generated continuations and the edited spans as targets. Offline applies supervision only to localized edits. This comparison tests the benefit of accurate credit assignment with the training data held fixed. Online additionally adapts the training data to the model's evolving failure distribution.

\paragraph{Relation to online imitation learning.}
Because edits can be generated from the current policy’s failed completions, the online version resembles DAgger-style imitation learning~\cite{ross2011dagger}. The editor does not provide a full expert trajectory from the initial state. Instead, the editor labels the states actually reached by the learner. In our setting, these states are failure prefixes in generated reasoning traces. This distinction is important because reasoning failures are often policy-dependent. A model may fail in ways that are specific to its own style, knowledge, and decoding behavior. Online edit training targets precisely these failures.
The online version is therefore a form of online corrective imitation: the learner is trained on the errors it actually induces, rather than on idealized demonstrations sampled from a fixed expert distribution.

\paragraph{Relation to trajectory-level RL.}
Similarly to reinforcement learning, our method does not require full expert demonstrations. The key difference is the form of feedback. Trajectory-level RL typically assigns a binary or scalar reward (or an advantage derived from such rewards) to an entire sampled completion. This indicates whether the completion succeeded but not what local change would have repaired it. Edit feedback provides this missing credit assignment. This localization is especially useful for long reasoning traces. A trajectory-level advantage may affect many tokens unrelated to the actual failure. Edit supervision instead concentrates the update at the editor-identified failure location, leaving the rest of the trajectory unsupervised. This can reduce credit diffusion and provide a lower-variance corrective signal.
Edit supervision may be especially useful early in training, when fully successful trajectories are rare. In that regime, reward-based methods may observe little useful variation, since many completions receive the same failing reward. Edits, by contrast, are available precisely on failed samples, making supervision abundant when the model most needs corrective feedback.
More broadly, edits and trajectory-level RL are complementary rather than substitutes. Edit supervision provides a localized textual specification of the corrected continuation that trajectory-level RL lacks at this granularity, while RL provides an outcome-grounded signal that edit supervision does not. Combining the two is a natural extension: edits to bootstrap the policy where successful trajectories are rare, RL to optimize for downstream task correctness once instruction compliance becomes routine.

\section{Related work}

\paragraph{Reasoning instruction following.}
ReasonIF~\citep{kwon2025reasonif} introduces a benchmark for evaluating whether reasoning LLMs follow user-specified instructions throughout their reasoning traces (multilingual reasoning, formatting, length control), and shows that compliance degrades as task difficulty increases. MathIF~\citep{fu2025mathif} similarly finds that instruction following degrades as reasoning ability scales in mathematical reasoning models.

\paragraph{Process supervision and step-level credit assignment.}
A large body of work attacks credit assignment in reasoning by densifying supervision at the step level. Process Reward Models~\citep{lightman2023letsverify,setlur2025rewardingprogress,zhou2025prmsurvey} train step-level critics from per-step annotations; OmegaPRM~\citep{luo2024omegaprm} automates first-error localization via Monte Carlo rollouts and divide-and-conquer MCTS to scale PRM training; and ELPO~\citep{elpo2026} localizes the first irrecoverable step in tool-integrated rollouts and concentrates GRPO advantage there. More directly tied to first-error detection, VPPO~\citep{liu2026vppo} uses a PRM to partition an incorrect rollout into a verified good prefix and an erroneous suffix and rewards the former while penalizing the latter, while CAPO~\citep{xie2025capo} uses an LLM-based generative PRM to convert per-step critiques into token-level rewards. All of these methods specify \emph{where} credit should be concentrated, but reweight rewards on the model's own (uncorrected) tokens; none specifies \emph{what} the model should have produced instead. Our editor closes this gap by supplying a concrete textual repair, answering the counterfactual question of what the correct continuation at the failure point would have been.

\paragraph{Localized SFT and token-level interventions.}
Closer in form to our objective, Critical Token Fine-Tuning (CFT)~\citep{ruan2025cft} also restricts the SFT loss to a sparse subset of token positions within a model-sampled trajectory, but operates on \emph{correct} rollouts and identifies positions via counterfactual perturbation, supervising the model's own original token at each chosen position. The criterion is tied to outcome-verifiable tasks and discards questions the model cannot already solve. \citet{lin2025critical} similarly show that replacing a single \emph{critical token} can flip an incorrect trace to a correct one, providing empirical evidence for our premise that small local interventions redirect downstream reasoning, and use a contrastive signal between models fine-tuned on correct and incorrect trajectories as a token-level weight inside DPO. In contrast, our method draws supervision from the model's \emph{failed} rollouts, requires no auxiliary contrastive models, and supervises editor-supplied replacement tokens at the points where the trajectory first violates the reasoning instruction.

\paragraph{Edit- and intervention-based supervision.}
Closest in form to our method are approaches that supply textual corrections at the failure point rather than scalar signals. SuperCorrect~\citep{yang2025supercorrect} shares the motivation that errors are localized and supervision should target the first failing step, but it trains a two-pass self-correction loop in which the student produces a trace, re-examines it, and emits a structured correction artifact, with the teacher supervising this corrective second pass via cross-model DPO.
The concurrent Intervention Training (InT) of \citet{yang2026int} is most closely related: given an incorrect rollout, InT self-verifies each step against a reference solution, locates the first erroneous step, and applies SFT to the prefix concatenated with a single-step intervention. We differ in three respects. First, InT targets final-answer accuracy and depends on reference solutions to verify steps; we target reasoning-process control, where no reference trace of the desired reasoning exists and the editor instead detects violations of the user-specified instruction.
Second, InT inserts a single intervention per trajectory, whereas we may apply successive edits, repeatedly identifying the next first failure until the trace complies.
Third, InT runs as a single offline pass; we show that an online variant, with edits continuously regenerated from the current policy's failure distribution, outperforms the offline procedure in domain.

\section{Experiments}

\subsection{Experimental setup}

The experiments instantiate the iterative edit procedure described in Section~\ref{subsec:method}. For each sampled trajectory, the editor locates the first remaining violation of the reasoning instruction, applies a minimal local replacement, and generation resumes from the corrected prefix. We repeat this process until the trajectory is compliant; if the stopping criterion is reached first, we discard the example.

Training prompts are constructed from the ReasonIF data-generation pipeline. The underlying tasks come from the same source datasets as ReasonIF: AIME~\citep{aimo2025aime}, AMC~\citep{aimo2025amc}, ARC~\cite{clark2018think}, GPQA~\citep{rein2023gpqa}, and GSM8K~\citep{cobbe2021training}. Each prompt is paired with one reasoning-process instruction from the six ReasonIF categories: Uppercase only, JSON formatting, Multilinguality, Word limit, Remove commas, and Disclaimer. These categories are intended as controlled tests of reasoning instruction following rather than as claims about optimal reasoning strategies or final-answer improvement. We choose training prompts that do not overlap with the ReasonIF benchmark test set.

\begin{table}[t]
\caption{Overview of the main results (higher is better). The first three rows report overall reasoning instruction-following accuracy (\%) in three settings: in-domain, on out-of-domain prompts, and under unseen reasoning instructions. The last two rows measure preservation of capabilities not targeted by training: final-answer accuracy (\%) on the ReasonIF tasks and on IMO-AnswerBench, both evaluated with the reasoning instruction removed. Trained systems report mean $\pm$ standard deviation across three training runs; among them, the best result in each row is shown in bold, together with any result whose standard deviation covers it. Per-category breakdowns of the instruction-following results are given in Tables~\ref{tab:reasonif-results}, \ref{tab:reasonif-hf-thinking-results}, and~\ref{tab:reasonif-ood-instructions-results}.}
\centering
\small
\begin{tabular*}{\textwidth}{@{\extracolsep{\fill}}lcccccc}
\toprule
Evaluation & Base & SFT & SFT-GPT & GRPO & Offline & Online \\
\midrule
ReasonIF (in-domain) & 13 & 40.3 $\pm$ 0.4 & 92 $\pm$ 2 & 64 $\pm$ 6 & 87 $\pm$ 1 & \textbf{97 $\pm$ 1} \\
Out-of-domain prompts & 21 & 57.1 $\pm$ 0.2 & \textbf{96.8 $\pm$ 0.4} & 80 $\pm$ 3 & \textbf{96.6 $\pm$ 0.2} & \textbf{95 $\pm$ 2} \\
Unseen reasoning instructions & 2 & 5 $\pm$ 1 & 63 $\pm$ 2 & 30 $\pm$ 10 & \textbf{68.4 $\pm$ 0.4} & 34 $\pm$ 9 \\
\midrule
ReasonIF answer accuracy & 96 & \textbf{96.9 $\pm$ 0.3} & 94 $\pm$ 1 & 91 $\pm$ 2 & 95 $\pm$ 1 & 95 $\pm$ 1 \\
IMO-AnswerBench & 61 & \textbf{61 $\pm$ 1} & 43 $\pm$ 2 & \textbf{60 $\pm$ 2} & 57 $\pm$ 1 & \textbf{59 $\pm$ 2} \\
\bottomrule
\end{tabular*}
\label{tab:overview}
\end{table}

We compare six systems (Table~\ref{tab:overview}): Base, SFT, SFT-GPT, GRPO, Offline, and Online. Base is the starting checkpoint, Qwen3-32B~\citep{qwen3technicalreport}, from which all trained systems are initialized; for the in-domain comparison, we additionally train all systems from Nemotron-3-Nano-Omni-30B-A3B-Reasoning-BF16~\citep{nvidia2026nemotron3nanoomni} (Nemotron). SFT is the full-trajectory baseline of Section~\ref{subsec:method}, trained on the same edited trajectories as Offline. SFT-GPT instead uses complete GPT-5~\citep{openai2025gpt5} responses to the same reasoning-instruction tasks, with KL regularization toward the base model on a separate sample from Infinity-Instruct~\citep{li2025infinityinstruct}. Further details on both SFT baselines are provided in Appendix~\ref{app:sft-data}. GRPO is our trajectory-level reinforcement-learning baseline, trained with a ReasonIF compliance reward. Offline implements edit-based training with a fixed set of edited rollouts, whereas Online regenerates edits from the current policy during training. For both edit-based methods, supervision is applied only to localized correction targets.

For edit generation, we use a hybrid editor. Scripted verifiers handle Uppercase only, Remove commas, and the boundary logic for Word limit. The LLM editor, GPT-5, fully handles Multilinguality, Disclaimer, and JSON repairs that require rewriting existing JSON-like content. The editor is prompted to produce the minimal replacement span that fixes the detected violation. Further implementation details are provided in Appendix~\ref{app:editor-details}.

\subsection{Main results}

We evaluate each system along two axes: how reliably it follows reasoning instructions, and how well it preserves the base model's capabilities that the reasoning-control objective does not target. For the first axis we report \emph{reasoning instruction-following accuracy}, the fraction of responses whose reasoning trace satisfies the requested instruction, in three settings: (i) \emph{in-domain}, on the ReasonIF benchmark; (ii) \emph{out-of-domain prompts}, new prompts from HuggingFace H4 Multilingual-Thinking~\citep{huggingfaceh4_2025_multilingual_thinking} paired with the original ReasonIF instructions; and (iii) \emph{unseen reasoning instructions}, the ReasonIF tasks paired with three reasoning-process instructions not seen during training (Appendix~\ref{app:unseen-instructions}). For the second axis we remove the reasoning instruction at evaluation and measure accuracy on IMO-AnswerBench~\citep{luong-etal-2025-towards}, an independent set of 400 olympiad-level problems that is never used for training; because it lies outside the training distribution, it tests whether a system retains the base model's mathematical problem-solving skill. Since the benchmark ships no evaluation script, we use GPT-5.4~\citep{openai2026gpt54} to judge each response against its ground-truth answer, so that the judge is independent of all evaluated models. Table~\ref{tab:overview} summarizes both axes; per-category breakdowns for each setting are provided in Appendix~\ref{app:detailed-results}.

The central finding is that edit-based training combines strong compliance and transfer with limited loss on untargeted tasks. Online attains the strongest in-domain control, transfers well to new prompts, and stays close to the base model on IMO-AnswerBench, so its gains in reasoning control come without a significant loss of untargeted mathematical ability. Every baseline gives up one side of this trade-off. SFT-GPT reaches comparable in-domain compliance, but its mathematical accuracy drops sharply. We attribute this decline to the mismatch between GPT-5's reasoning trajectories and Qwen3's native reasoning: imitating complete teacher trajectories may alter behaviors that support the base model's performance. In line with this, SFT on Qwen3's \emph{own} trajectories with local edits retains mathematical accuracy, but achieves lower instruction compliance and weaker transfer to new prompts than the edit-based methods. GRPO with a binary compliance reward improves compliance far less than the edit-based methods, despite training directly on that reward. Offline's higher compliance than SFT highlights the benefits of accurate credit assignment: both use the same underlying trajectories, but Offline applies supervision only to localized edits instead of applying full-trajectory imitation. Offline edit-based training reaches lower in-domain compliance than Online but outperforms SFT and GRPO, and notably transfers best of all systems to unseen instructions, while giving up a little mathematical accuracy. Both variants start from the same failure distribution, so the difference likely arises during training: once the first violations are repaired, Online's regenerated edits concentrate on the residual violations within the six training categories, whereas Offline retains the base model's full initial failure distribution. Consistent with this, Online's deficit is specific to unseen instruction structures and not to new prompts.

A further preservation check supports this picture. With the reasoning instruction removed, final-answer correctness on the ReasonIF tasks remains near the base result for all systems but GRPO (Table~\ref{tab:overview}); but these tasks come from sources seen during training and cannot reveal skill loss outside that distribution, which is exactly what IMO-AnswerBench exposes for SFT-GPT. Across these Qwen3-32B evaluations, the edit-based methods pair strong, transferable instruction following with largely preserved capability: Online gives the strongest in-domain control, while Offline generalizes best to unseen instructions.

The same in-domain conclusions hold for Nemotron, a model from a different family: Online again reaches the highest compliance, and Offline outperforms SFT and GRPO (Table~\ref{tab:nemotron-reasonif-results} in Appendix~\ref{app:detailed-results}).

Finally, removing the KL regularizer that the edit-based methods apply to the generated continuations (Section~\ref{subsec:method}) retains most of Offline's in-domain compliance, but substantially increases variability across training runs and reduces transfer to unseen instructions; even without regularization, Offline still outperforms SFT in all three settings (Appendix~\ref{app:regularization-ablation}).

\begin{table}[t]
\caption{IFEval constraint accuracy (\%), macro-averaged over constraint categories. Online$^{-}$ and Online$^{+}$ evaluate the edit-trained model without and with the draft-review-revise reasoning instruction, respectively; Base is evaluated without it. \emph{Seen} comprises the three constraint categories represented during training and \emph{Unseen} the remaining categories; \emph{Constraints} indicates whether the prompt imposes a single constraint or multiple.}
\centering
\small
\begin{tabular*}{\textwidth}{@{\extracolsep{\fill}}llccc}
\toprule
Categories & Constraints & Base & Online$^{-}$ & Online$^{+}$ \\
\midrule
Seen & Single & 85 & 89 & 96 \\
Seen & Multiple & 81 & 85 & 85 \\
Unseen & Single & 93 & 93 & 91 \\
Unseen & Multiple & 90 & 89 & 88 \\
\bottomrule
\end{tabular*}
\label{tab:ifeval-transfer}
\end{table}

\subsection{From reasoning control to task performance}
\label{subsec:ifeval}

The reasoning instructions studied so far are controlled proxies. We now ask whether teaching a reasoning procedure can improve the final answer itself, using IFEval~\citep{zhou2023instruction}, whose prompts impose an output constraint on the response, such as a forbidden word or a capitalization rule. Each training prompt pairs one constraint, drawn from Capital Word Frequency, English Capital, or Forbidden Words, with a reasoning instruction that asks the model to draft a response, review it against the constraint, and revise it when the review finds a violation. Every training prompt imposes exactly one output constraint, so prompts carrying several constraints at once are never seen during training; we therefore report the seen and unseen categories separately for single- and multiple-constraint prompts. We teach this draft-review-revise procedure with edit-based training (the online variant).

Crucially, the editor does more than enforce the format of this procedure. Whenever the procedure breaks down, it restores the draft-review-revise structure, but it also corrects its content: it fixes reviews that miss or misidentify a violation, repairs unresolved violations, and forces the final response to reproduce the corrected draft. Only editor-changed spans are supervised. This supervision targets both execution of the procedure and the correctness of its review and revision steps.

When instructed, Base never completes the draft-review-revise procedure, which is why it is evaluated without the instruction; Online follows it in 92\% of responses. Table~\ref{tab:ifeval-transfer} reports macro-averaged IFEval constraint accuracy. On the constraint categories seen during training, accuracy improves over Base, most clearly when the reasoning instruction is provided at test time. On single-constraint prompts, training alone yields a non-significant gain of 4 percentage points ($\pm$5; 95\% confidence interval over prompts, paired across systems); invoking the procedure adds 7 ($\pm$6), for a total gain of 11 ($\pm$8). On multiple-constraint prompts, accuracy on seen constraint categories improves marginally, from 81\% to 85\% ($\pm$5), despite training exclusively on single-constraint prompts. On unseen constraint categories, the effects are mixed: compliance improves for some and declines for others, with a slight, non-significant decrease on average when the procedure is invoked. This variation may reflect differences in the model's ability to detect violations of different constraints. Detection remains the main bottleneck overall: of 631 noncompliant first drafts, the model flags and revises only 18\%, and 62\% of those revisions succeed. Thus, correction is often effective once a problem is identified, but most violations go undetected.

\section{Conclusion}

We introduced edit-based training, a framework that learns from a model's own reasoning traces by correcting them locally rather than imitating or reweighting them as a whole. An editor identifies the first point at which a trace violates a requested reasoning instruction and proposes a minimal replacement span; only these edited spans are supervised. This concentrates the training signal exactly where the model goes wrong. By restricting corrective supervision to short spans within the model's own trajectories, the method aims to limit changes unrelated to the requested reasoning instruction. We instantiated the framework in offline and online variants for reasoning instruction following. Edit-based training improves in-domain compliance on two distinct model families, Qwen3-32B and Nemotron. On Qwen3-32B, where we evaluate further, the gains extend to out-of-domain prompts and unseen instructions while largely preserving the model's skills on the untargeted tasks we measure, unlike full-trajectory distillation, which attains strong compliance only at the cost of untargeted capabilities, and trajectory-level reinforcement learning, which improves compliance far less. Localized correction is thus an effective complement to full-trajectory imitation and reinforcement learning for shaping how language models reason.

We see two main directions for future work. First, our study uses reasoning-instruction compliance as a controlled proxy; a natural next step is to apply edit-based training to improve primary task performance directly, with an editor that corrects substantive reasoning errors rather than instruction violations. Second, the LLM editor here is an external frontier model. As open models improve, the same framework could enable self-improvement, in which a model edits its own thinking and trains on its own corrections, removing the dependence on a stronger external editor.

\paragraph{Limitations.} The six ReasonIF categories are deliberately verifiable proxies and do not span the full range of reasoning controllability needs; however, the IFEval experiment in Section~\ref{subsec:ifeval} tests a substantive reasoning procedure, in which the model first drafts a response and then verifies the draft against the requirements. This draft-review-revise procedure was trained only in the Online regime, and transfer from it is concentrated in the constraint categories represented during training; it does not extend to the remaining ones, where invoking the procedure gives a slight, non-significant decrease in accuracy. The edit dataset was generated once rather than regenerated per seed, leaving variance contributed by the LLM editor unmeasured. Nemotron was evaluated in domain only; all transfer, downstream, and preservation results use Qwen3-32B. The DPO alternative formulated in Appendix~\ref{app:dpo-loss} is left to future work.

\begin{ack}
We acknowledge CSC -- IT Center for Science, Finland, for awarding this project access to the LUMI supercomputer, owned by the EuroHPC Joint Undertaking, hosted by CSC (Finland) and the LUMI consortium through CSC.
We used an AI assistant for writing support (wording, consistency checks, LaTeX layout, illustrations) and for drafting parts of the analysis scripts; all outputs were verified by the authors.
The authors declare no competing interests.
\end{ack}

\bibliographystyle{unsrtnat}
\bibliography{refs}

\newpage
\appendix

\section{Detailed evaluation results}
\label{app:detailed-results}

This appendix reports the per-category breakdowns of the reasoning instruction-following accuracy summarized in Table~\ref{tab:overview}. Tables~\ref{tab:reasonif-results} and~\ref{tab:nemotron-reasonif-results} give the in-domain results for Qwen3-32B and Nemotron, respectively. Tables~\ref{tab:reasonif-hf-thinking-results} and~\ref{tab:reasonif-ood-instructions-results} give the Qwen3-32B results for out-of-domain prompts and unseen instructions. Trained systems report mean $\pm$ standard deviation across three training runs; per-sample evaluation uncertainty is approximately binomial and is omitted from the tables. The unseen-instruction setting (Table~\ref{tab:reasonif-ood-instructions-results}) uses three reasoning-process instructions held out from training. \emph{Answer draft} asks the model to include an explicit answer draft inside delimiter tags while reasoning; \emph{Sections} asks it to organize its reasoning into a fixed sequence of labeled parts (question analysis, answer draft, answer verification, and final answer); and \emph{Two approaches} asks it to compare two candidate solution approaches before selecting one. These differ from the six training categories in both form and intent, testing whether a trained system follows reasoning instructions with an unfamiliar structure; their exact wording is given in Appendix~\ref{app:unseen-instructions}.

\begin{table}[h]
\caption{Qwen3-32B reasoning instruction-following accuracy (\%) on the ReasonIF benchmark. Each training run is evaluated six times; 1800 responses per training run.}
\centering
\small
\begin{tabular*}{\textwidth}{@{\extracolsep{\fill}}lcccccc}
\toprule
Category & Base & SFT & SFT-GPT & GRPO & Offline & Online \\
\midrule
Uppercase only & 0 & 33 $\pm$ 3 & 93 $\pm$ 2 & 38 $\pm$ 29 & 70 $\pm$ 2 & 96 $\pm$ 1 \\
JSON formatting & 0 & 66 $\pm$ 1 & 91 $\pm$ 8 & 0 $\pm$ 0 & 75 $\pm$ 1 & 95 $\pm$ 1 \\
Multilinguality & 44 & 78 $\pm$ 2 & 94.0 $\pm$ 0.3 & 100 $\pm$ 0 & 98 $\pm$ 2 & 99 $\pm$ 1 \\
Word limit & 0 & 7 $\pm$ 2 & 94 $\pm$ 2 & 65 $\pm$ 14 & 99.9 $\pm$ 0.1 & 99 $\pm$ 1 \\
Remove commas & 26 & 48 $\pm$ 2 & 99.0 $\pm$ 0.2 & 97 $\pm$ 2 & 94 $\pm$ 1 & 99 $\pm$ 1 \\
Disclaimer & 7 & 9 $\pm$ 1 & 83 $\pm$ 3 & 77 $\pm$ 2 & 79 $\pm$ 2 & 94 $\pm$ 4 \\
\midrule
Overall & 13 & 40.3 $\pm$ 0.4 & 92 $\pm$ 2 & 64 $\pm$ 6 & 87 $\pm$ 1 & 97 $\pm$ 1 \\
\bottomrule
\end{tabular*}
\label{tab:reasonif-results}
\end{table}

\begin{table}[h]
\caption{Nemotron-3-Nano-Omni-30B-A3B-Reasoning-BF16 reasoning instruction-following accuracy (\%) on the ReasonIF benchmark. Each training run is evaluated six times; 1800 responses per training run.}
\centering
\small
\begin{tabular*}{\textwidth}{@{\extracolsep{\fill}}lcccccc}
\toprule
Category & Base & SFT & SFT-GPT & GRPO & Offline & Online \\
\midrule
Uppercase only & 0 & 36 $\pm$ 1 & 90 $\pm$ 2 & 0 $\pm$ 0 & 29 $\pm$ 3 & 91 $\pm$ 1 \\
JSON formatting & 0 & 83 $\pm$ 6 & 91 $\pm$ 2 & 0 $\pm$ 0 & 90 $\pm$ 3 & 99.5 $\pm$ 0.3 \\
Multilinguality & 27 & 73 $\pm$ 1 & 93 $\pm$ 1 & 27.1 $\pm$ 0.3 & 92 $\pm$ 2 & 94.6 $\pm$ 0.5 \\
Word limit & 13 & 16 $\pm$ 2 & 89 $\pm$ 1 & 8 $\pm$ 6 & 87 $\pm$ 7 & 93 $\pm$ 2 \\
Remove commas & 4 & 46 $\pm$ 1 & 97.7 $\pm$ 0.4 & 36 $\pm$ 3 & 97 $\pm$ 1 & 100 $\pm$ 0 \\
Disclaimer & 3 & 45 $\pm$ 2 & 78 $\pm$ 2 & 2 $\pm$ 2 & 71 $\pm$ 18 & 97 $\pm$ 0 \\
\midrule
Overall & 8 & 50 $\pm$ 1 & 89.9 $\pm$ 0.4 & 13 $\pm$ 2 & 79 $\pm$ 1 & 96 $\pm$ 1 \\
\bottomrule
\end{tabular*}
\label{tab:nemotron-reasonif-results}
\end{table}

\begin{table}[h]
\caption{Qwen3-32B reasoning instruction-following accuracy (\%) on a dataset constructed from the HuggingFace H4 Multilingual-Thinking dataset~\citep{huggingfaceh4_2025_multilingual_thinking} with ReasonIF instructions appended. Each training run is evaluated twice; 2000 responses per training run.}
\centering
\small
\begin{tabular*}{\textwidth}{@{\extracolsep{\fill}}lcccccc}
\toprule
Category & Base & SFT & SFT-GPT & GRPO & Offline & Online \\
\midrule
Uppercase only & 0 & 83 $\pm$ 1 & 94 $\pm$ 1 & 83 $\pm$ 16 & 91.2 $\pm$ 0.4 & 94 $\pm$ 3 \\
JSON formatting & 0 & 6 $\pm$ 3 & 94 $\pm$ 3 & 0 $\pm$ 0 & 94.1 $\pm$ 0.8 & 96 $\pm$ 3 \\
Multilinguality & 5 & 33 $\pm$ 2 & 99.3 $\pm$ 0.3 & 99.7 $\pm$ 0.3 & 98.1 $\pm$ 0.6 & 93 $\pm$ 1 \\
Word limit & 81 & 85 $\pm$ 1 & 99.1 $\pm$ 0.1 & 99.7 $\pm$ 0.5 & 99.1 $\pm$ 0.3 & 91 $\pm$ 5 \\
Remove commas & 16 & 87 $\pm$ 2 & 97 $\pm$ 1 & 100 $\pm$ 0 & 100 $\pm$ 0 & 99.7 $\pm$ 0.1 \\
Disclaimer & 22 & 48 $\pm$ 1 & 98 $\pm$ 1 & 98 $\pm$ 1 & 97 $\pm$ 1 & 98.8 $\pm$ 0.6 \\
\midrule
Overall & 21 & 57.1 $\pm$ 0.2 & 96.8 $\pm$ 0.4 & 80 $\pm$ 3 & 96.6 $\pm$ 0.2 & 95 $\pm$ 2 \\
\bottomrule
\end{tabular*}
\label{tab:reasonif-hf-thinking-results}
\end{table}

\begin{table}[h]
\caption{Qwen3-32B reasoning instruction-following accuracy (\%) on the ReasonIF benchmark with the original instructions replaced by unseen reasoning-process instructions. Each training run is evaluated three times; 900 responses per training run.}
\centering
\small
\begin{tabular*}{\textwidth}{@{\extracolsep{\fill}}lcccccc}
\toprule
Category & Base & SFT & SFT-GPT & GRPO & Offline & Online \\
\midrule
Answer draft & 1 & 0.9 $\pm$ 0.4 & 43 $\pm$ 5 & 34 $\pm$ 7 & 60 $\pm$ 1 & 33 $\pm$ 14 \\
Sections & 2 & 8 $\pm$ 2 & 63 $\pm$ 1 & 24 $\pm$ 10 & 53 $\pm$ 2 & 30 $\pm$ 9 \\
Two approaches & 4 & 6 $\pm$ 1 & 82 $\pm$ 1 & 30 $\pm$ 13 & 89 $\pm$ 2 & 38 $\pm$ 5 \\
\midrule
Overall & 2 & 5 $\pm$ 1 & 63 $\pm$ 2 & 30 $\pm$ 10 & 68.4 $\pm$ 0.4 & 34 $\pm$ 9 \\
\bottomrule
\end{tabular*}
\label{tab:reasonif-ood-instructions-results}
\end{table}

\subsection{Effect of regularization}
\label{app:regularization-ablation}

We evaluate Qwen3-32B trained with Offline without the regularization described in Appendix~\ref{app:training-configuration}. We run this ablation on Offline only: its fixed edit dataset isolates the effect of regularization from changes in the training data. Table~\ref{tab:regularization-ablation} compares this ablation with the regularized Offline results from Table~\ref{tab:overview}.

\begin{table}[h]
\caption{Qwen3-32B Offline reasoning instruction-following accuracy (\%) with and without regularization. Both settings report mean $\pm$ standard deviation across three training runs.}
\centering
\small
\begin{tabular}{lcc}
\toprule
Evaluation & With reg. & Without reg. \\
\midrule
ReasonIF (in-domain) & 87 $\pm$ 1 & 85 $\pm$ 2 \\
Out-of-domain prompts & 96.6 $\pm$ 0.2 & 94 $\pm$ 1 \\
Unseen reasoning instructions & 68.4 $\pm$ 0.4 & 30 $\pm$ 8 \\
\bottomrule
\end{tabular}
\label{tab:regularization-ablation}
\end{table}

Removing regularization increases variability across training runs in all three settings, suggesting that regularization supports training stability. Mean compliance declines modestly in domain and on new prompts, but sharply on unseen instructions. This increased instability may partly explain the decline in generalization to unseen instruction structures.

\section{Experimental details}
\label{app:experimental-details}

\subsection{Prompt construction}
\label{app:prompt-construction}

Each raw task prompt is wrapped in a fixed template that presents the task statement, inserts the reasoning-process instruction into a delimited thinking-guidelines block, and asks the model to place the final answer inside answer tags:

\begin{verbatim}
{question_statement}

<thinking guidelines>
{constraint_text}
</thinking guidelines>

Place the final answer inside the tags `<answer>` and `</answer>`.
\end{verbatim}

Here, \texttt{question\_statement} is the underlying task prompt and \texttt{constraint\_text} is the natural-language reasoning-process instruction for that example.

\subsection{Unseen reasoning instructions}
\label{app:unseen-instructions}

For the transfer evaluation with unseen reasoning-process instructions, we keep the ReasonIF task prompts and replace the original instruction categories with the following three instructions.

\paragraph{Answer draft.}
\begin{verbatim}
When reasoning, provide the answer draft between an '<answer_draft>' tag
and a '</answer_draft>' tag. Put the '<answer_draft>' tag on its own line,
and put the '</answer_draft>' tag on its own line.
\end{verbatim}

\paragraph{Sections.}
\begin{verbatim}
When reasoning, include these exact labeled parts in this order:

SECTION: QUESTION ANALYSIS
SECTION: ANSWER DRAFT
SECTION: ANSWER VERIFICATION
SECTION: FINAL ANSWER
\end{verbatim}

\paragraph{Two approaches.}
\begin{verbatim}
When reasoning, include a structured comparison of two possible solution
approaches and then choose one. Use these exact labels in this order:

Question Analysis:
Approach 1:
Approach 2:
Approach Selection:
\end{verbatim}

\subsection{IFEval editor and prompt construction}
\label{app:ifeval-transfer}

Training and validation prompts (3k and 300) for the IFEval experiment of Section~\ref{subsec:ifeval} are generated with GPT-5.4~\citep{openai2026gpt54}; the original IFEval set (541 prompts) is used only for testing. Each prompt is appended with one IFEval instruction, selected from the three training categories (Capital Word Frequency, English Capital, and Forbidden Words), and with the thinking guidelines given below, which ask the model to draft a response, review it against the instruction, and revise it when the review finds a violation. GPT-5.4 also serves as the editor: it edits the trajectory whenever the draft-review-revise procedure breaks down. It restores the draft-review-revise structure, corrects inaccurate reviews or unresolved problems in the revised draft, and makes the final response reproduce the revised draft byte for byte. Only spans changed by the editor contribute to the training loss. This experiment uses a single training run; each system is evaluated six times, and Table~\ref{tab:ifeval-transfer} reports the average over these evaluations.

The reasoning instruction for this experiment is:

\begin{Verbatim}[breaklines=true,breakautoindent=false,breaksymbolleft={},breaksymbolright={}]
First reason normally toward an answer candidate. Then, while reasoning, use the XML section structure below. Put each opening tag and closing tag on its own line.

<answer_candidate>
Draft the answer candidate. This should be the response you would show to the user if it passes review.
</answer_candidate>

<answer_candidate_review>
Review the answer candidate for correctness and compliance with the user's constraints using this format:

Requirement 1: [requirement name]. [reasoning about compliance, including specific evidence or an argument from checking the answer_candidate] — PASS or FAIL
Requirement 2: [requirement name]. [reasoning about compliance, including specific evidence or an argument from checking the answer_candidate] — PASS or FAIL
Continue for each explicit requirement or constraint.

Each requirement line must name the requirement and reason about whether the answer_candidate satisfies it. The reasoning must include specific evidence or an argument from checking the answer_candidate, not a generic statement.

Verdict: PASS or FAIL

If the verdict is FAIL, explain briefly how to fix the failed requirements.
</answer_candidate_review>

If the review verdict is PASS, finish reasoning and output the answer_candidate as the final answer.

If the review verdict is FAIL, reason normally about how to fix each failed requirement. Then include the optional corrected final candidate section:

<improved_answer_candidate>
Write the corrected final answer. It must fix the failed requirements while preserving the parts of the answer_candidate that were already correct.
</improved_answer_candidate>
\end{Verbatim}

\subsection{Full-trajectory SFT baselines}
\label{app:sft-data}

For SFT, we construct a full-trajectory dataset from the same edited rollouts used by Offline. Both the model's own generated continuations and the editor's replacement spans serve as supervised targets. SFT applies cross-entropy over the full trajectories without a separate KL term. Most of its targets are continuations sampled from the base model, so this loss acts as a sample-based regularizer toward the base model, related to, though not identical with, the KL term used by the edit-based methods.

SFT-GPT instead uses complete compliant responses generated by GPT-5 for the same reasoning-instruction tasks and applies supervised fine-tuning to the full trajectories. It therefore measures full-trajectory distillation from a stronger teacher, while SFT uses the target model's own trajectories with local edits.

To reduce general distributional drift, we regularize SFT-GPT with a KL divergence to the base model on a separate sample from Infinity-Instruct~\citep{li2025infinityinstruct}.

\subsection{Editor implementation}
\label{app:editor-details}

The editor combines scripted verifiers with an LLM editor, GPT-5. The scripted verifiers implement the direct repairs used for Uppercase only, Remove commas, Word limit boundary cases, and JSON scaffolding. The LLM editor operates at the character level and is used for Multilinguality, Disclaimer, and JSON repair when the reasoning already contains JSON-like content. Its prompt asks the model to identify the earliest character position at which the response violates the constraint, keep all preceding text byte-identical, and edit only the suffix needed to make the response compliant while preserving fluency and meaning except where the constraint explicitly requires a change. The corrected span is then extracted from this suffix as described in Section~\ref{subsec:method}. We allow at most 10 edit iterations per trajectory; examples that remain non-compliant after this limit are discarded.

For Uppercase only and Remove commas, the script directly rewrites the surface form. For Word limit, it acts at the reasoning/answer boundary: if the limit is exceeded, we stop the reasoning and force the answer component to begin; if the reasoning uses less than one fifth of the allowed quota, we force a reasoning continuation by querying the model again with the task prompt and the reasoning prefix, but without the word-limit guideline. For JSON formatting, we insert markdown formatting and an opening curly brace when no JSON-like content is present; otherwise GPT-5 repairs malformed JSON-like reasoning. Multilinguality and Disclaimer are handled by GPT-5.

\subsection{Existing assets}
\label{app:existing-assets}

We use existing public assets for the base models, instruction benchmarks, underlying task prompts, and regularization data. Qwen3-32B~\citep{qwen3technicalreport}, ReasonIF~\citep{kwon2025reasonif}, IFEval~\citep{zhou2023instruction}, HuggingFace H4 Multilingual-Thinking~\citep{huggingfaceh4_2025_multilingual_thinking}, and the AI-MO AIME/AMC sets~\citep{aimo2025aime,aimo2025amc} are released under Apache-2.0. Nemotron-3-Nano-Omni-30B-A3B-Reasoning-BF16~\citep{nvidia2026nemotron3nanoomni} is released under the NVIDIA Open Model Agreement. The other datasets are released under MIT for GSM8K~\citep{cobbe2021training}, CC BY 4.0 for GPQA~\citep{rein2023gpqa} and IMO-AnswerBench~\citep{luong-etal-2025-towards,hwilner2025imoanswerbench}, and CC BY-SA 4.0 for AI2 ARC~\cite{clark2018think} and Infinity-Instruct~\citep{li2025infinityinstruct}. We use these assets only for research training and evaluation and do not redistribute the original datasets.

\subsection{Training configuration}
\label{app:training-configuration}

All trained systems fine-tune Qwen3-32B or Nemotron-3-Nano-Omni-30B-A3B-Reasoning-BF16 using LoRA. The LoRA rank is 64, LoRA alpha is 128, and LoRA dropout is 0.05. LoRA adapters are applied to all linear transformer modules except the final language-modeling head. Optimization uses AdamW with weight decay set to zero and learning rate $2\times10^{-5}$. Offline, SFT, and SFT-GPT use linear learning-rate decay; the online methods, Online and GRPO, use no decay.

The edit-based training loss is cross-entropy on the corrected tokens. We regularize the edit-based methods with a KL divergence to the base model on the model continuations from the same training examples, i.e., the tokens generated after the local correction; these continuation tokens are excluded from the primary loss. The two terms are weighted 1:1, and each is averaged over its own target tokens. For GRPO, we use the standard KL regularization term with coefficient 0.05.

Online training uses a buffer of generated samples. Batch construction prioritizes fresh unused samples from the most recent model checkpoint, then other unused samples, and finally samples drawn uniformly from the full buffer if the batch is still not full.

Rollouts for both model families use temperature 0.6 and top-$p$ 0.95, with no presence, frequency, or repetition penalties. Qwen3 additionally uses top-$k$ 20; Nemotron uses no top-$k$ truncation. A representative Qwen3 run uses about 700--850 gradient steps and takes about 10 hours of wall-clock time.

Experiments are run on the LUMI supercomputer using the standard-g GPU partition. Each GPU node contains four AMD Instinct MI250X accelerators (exposed as eight logical GPUs via dual GCDs), a 64-core AMD EPYC 7A53 CPU, 512 GB system memory, and access to a high-performance parallel Lustre storage system. Each MI250X provides 128 GB of HBM2e memory. Each run uses two nodes: one node for training and one node for rollout generation. Training uses PyTorch FSDP2 across the GPUs of the training node.

\section{Loss function alternative}
\label{app:dpo-loss}

Section~\ref{subsec:method} introduces the cross-entropy edit loss in (\ref{eq:editloss}) and notes Direct Preference Optimization~\citep{rafailov2023direct} as a natural alternative. Treating each edit as a preference pair at the failure prefix $s_k$, with the editor's replacement $y_k^\star$ as the preferred continuation and the model's original span $y_k^{\mathrm{orig}}$ as the dispreferred one, the corresponding DPO objective is
\begin{equation}
\ell_{\mathrm{DPO}}(\pi)
=
- \mathbb{E}_{s_k \sim D_{\mathrm{fail}}^\pi}
\left[
\log \sigma\!\left(
\beta \log \frac{\pi(y_k^\star \mid s_k)}{\pi_{\mathrm{ref}}(y_k^\star \mid s_k)}
- \beta \log \frac{\pi(y_k^{\mathrm{orig}} \mid s_k)}{\pi_{\mathrm{ref}}(y_k^{\mathrm{orig}} \mid s_k)}
\right)
\right],
\end{equation}
where $\pi_{\mathrm{ref}}$ is a frozen reference policy and $\beta$ controls the strength of the implicit KL regularization toward it. Compared with the cross-entropy formulation, this would explicitly push probability mass away from the original failure continuation rather than only toward the corrected span. We leave a systematic comparison to future work.

\section{Broader impacts}
\label{app:broader-impacts}

Edit-based training is a method for making language models more controllable in their reasoning, which we view as a net-positive direction for alignment: instructions that the model reliably follows can be used to require auditable structure, block known shortcut patterns, or enforce verification steps that the model would otherwise skip. The same controllability could in principle be used to enforce reasoning structures that are themselves undesirable, for example by suppressing safety-relevant intermediate reasoning. Doing so requires the operator to specify the constraint explicitly and to obtain edit supervision that exemplifies it; both of these are inspectable artifacts rather than latent model behaviors. We do not release a new pretrained model, dataset, or scraped resource, so the work does not introduce additional release-time misuse vectors beyond those already present in the publicly available base model and benchmarks we use.

\newpage
\section*{NeurIPS Paper Checklist}

\begin{enumerate}

\item {\bf Claims}
    \item[] Question: Do the main claims made in the abstract and introduction accurately reflect the paper's contributions and scope?
    \item[] Answer: \answerYes{}.
    \item[] Justification: The abstract and introduction describe the edit-based method and claim improved reasoning-instruction compliance while largely preserving the base model's skills on the untargeted tasks we measure. These claims are supported by the results summarized in Table~\ref{tab:overview}, with per-setting breakdowns in Tables~\ref{tab:reasonif-results}, \ref{tab:nemotron-reasonif-results}, \ref{tab:reasonif-hf-thinking-results}, and \ref{tab:reasonif-ood-instructions-results}, and the regularization ablation in Appendix~\ref{app:regularization-ablation}.

\item {\bf Limitations}
    \item[] Question: Does the paper discuss the limitations of the work performed by the authors?
    \item[] Answer: \answerYes{}.
    \item[] Justification: A dedicated Limitations paragraph at the end of the Conclusion lists the main scope limits: the six ReasonIF categories cover only verifiable instances of reasoning controllability; the draft-review-revise procedure transfers only to constraint categories seen in training; the edit dataset is fixed across seeds, so editor-induced variance is unmeasured; Nemotron is evaluated in domain only, and all transfer, downstream, and preservation results use Qwen3-32B; and the DPO loss alternative is left to future work.

\item {\bf Theory assumptions and proofs}
    \item[] Question: For each theoretical result, does the paper provide the full set of assumptions and a complete (and correct) proof?
    \item[] Answer: \answerNA{}.
    \item[] Justification: The paper introduces a training objective and provides conceptual connections to online imitation learning and reinforcement learning, but it does not state formal theorems or proofs.

    \item {\bf Experimental result reproducibility}
    \item[] Question: Does the paper fully disclose all the information needed to reproduce the main experimental results of the paper to the extent that it affects the main claims and/or conclusions of the paper (regardless of whether the code and data are provided or not)?
    \item[] Answer: \answerYes{}.
    \item[] Justification: The experimental setup specifies the base model, datasets, compared systems, evaluation settings, number of runs, editor construction, and metrics. Additional implementation, SFT baseline construction, prompt, and training details are provided in Appendix~\ref{app:experimental-details}.

\item {\bf Open access to data and code}
    \item[] Question: Does the paper provide open access to the data and code, with sufficient instructions to faithfully reproduce the main experimental results, as described in supplemental material?
    \item[] Answer: \answerNo{}.
    \item[] Justification: The paper does not include code or data. The editor is described in detail in Appendix~\ref{app:editor-details}; together with the ReasonIF instruction code available on GitHub, this description allows the data-generation pipeline to be reproduced. Training hyperparameters are provided in Appendix~\ref{app:training-configuration}.

\item {\bf Experimental setting/details}
    \item[] Question: Does the paper specify all the training and test details (e.g., data splits, hyperparameters, how they were chosen, type of optimizer) necessary to understand the results?
    \item[] Answer: \answerYes{}.
    \item[] Justification: Section ``Experimental setup'' describes the train/test construction, evaluated settings, systems, metrics, and number of runs. Appendix~\ref{app:training-configuration} gives LoRA, optimizer, sampling, regularization, and online-buffer details.

\item {\bf Experiment statistical significance}
    \item[] Question: Does the paper report error bars suitably and correctly defined or other appropriate information about the statistical significance of the experiments?
    \item[] Answer: \answerYes{}.
    \item[] Justification: All ReasonIF tables report mean $\pm$ standard deviation across three independent training runs for SFT, SFT-GPT, GRPO, Offline, and Online. Base is the unmodified starting checkpoint and is reported as a single evaluation point. Per-sample evaluation uncertainty is approximately binomial and omitted from the tables, as stated in the captions and Appendix~\ref{app:detailed-results}. The IFEval experiment uses a single training run evaluated six times; Section~\ref{subsec:ifeval} reports paired 95\% confidence intervals over prompts (Appendix~\ref{app:ifeval-transfer}).

\item {\bf Experiments compute resources}
    \item[] Question: For each experiment, does the paper provide sufficient information on the computer resources (type of compute workers, memory, time of execution) needed to reproduce the experiments?
    \item[] Answer: \answerYes{}.
    \item[] Justification: Appendix~\ref{app:training-configuration} describes the LUMI standard-g nodes used for training and rollout generation, including accelerator type/count, CPU, memory, storage setup, distributed-training scheme, and approximate wall-clock and node-hour cost per representative run.

\item {\bf Code of ethics}
    \item[] Question: Does the research conducted in the paper conform, in every respect, with the NeurIPS Code of Ethics \url{https://neurips.cc/public/EthicsGuidelines}?
    \item[] Answer: \answerYes{}.
    \item[] Justification: The work uses existing public benchmarks and model training/evaluation procedures, without human subjects or sensitive personal data collection. We believe the research conforms to the NeurIPS Code of Ethics.

\item {\bf Broader impacts}
    \item[] Question: Does the paper discuss both potential positive societal impacts and negative societal impacts of the work performed?
    \item[] Answer: \answerYes{}.
    \item[] Justification: Appendix~\ref{app:broader-impacts} discusses both directions: reasoning controllability as an alignment lever (auditable structure, blocking known shortcut patterns, enforcing verification steps) and the corresponding misuse direction in which the same controllability is turned against safety-relevant intermediate reasoning. The work does not release a new pretrained model, dataset, or scraped resource, so it does not introduce release-time misuse vectors beyond those already present in the publicly available base model and benchmarks we use.

\item {\bf Safeguards}
    \item[] Question: Does the paper describe safeguards that have been put in place for responsible release of data or models that have a high risk for misuse (e.g., pre-trained language models, image generators, or scraped datasets)?
    \item[] Answer: \answerNA{}.
    \item[] Justification: The paper does not announce the release of a new pretrained language model, high-risk dataset, or other asset requiring special release safeguards. It fine-tunes an existing model for research evaluation.

\item {\bf Licenses for existing assets}
    \item[] Question: Are the creators or original owners of assets (e.g., code, data, models), used in the paper, properly credited and are the license and terms of use explicitly mentioned and properly respected?
    \item[] Answer: \answerYes{}.
    \item[] Justification: The paper cites the existing models and datasets used, including Qwen3-32B, Nemotron-3-Nano-Omni-30B-A3B-Reasoning-BF16, ReasonIF, IFEval, AIME/AMC, ARC, GPQA, GSM8K, HuggingFace H4 Multilingual-Thinking, IMO-AnswerBench, and Infinity-Instruct. Appendix~\ref{app:existing-assets} summarizes the public license information for these assets.

\item {\bf New assets}
    \item[] Question: Are new assets introduced in the paper well documented and is the documentation provided alongside the assets?
    \item[] Answer: \answerNA{}.
    \item[] Justification: The submission does not currently release a new dataset, benchmark, model checkpoint, or code package as a standalone asset. The generated training examples and edited rollouts are used for the reported experiments.

\item {\bf Crowdsourcing and research with human subjects}
    \item[] Question: For crowdsourcing experiments and research with human subjects, does the paper include the full text of instructions given to participants and screenshots, if applicable, as well as details about compensation (if any)?
    \item[] Answer: \answerNA{}.
    \item[] Justification: The paper does not involve crowdsourcing experiments or research with human subjects. All experiments are model training and evaluation on existing task datasets.

\item {\bf Institutional review board (IRB) approvals or equivalent for research with human subjects}
    \item[] Question: Does the paper describe potential risks incurred by study participants, whether such risks were disclosed to the subjects, and whether Institutional Review Board (IRB) approvals (or an equivalent approval/review based on the requirements of your country or institution) were obtained?
    \item[] Answer: \answerNA{}.
    \item[] Justification: No human-subject or crowdsourcing study was conducted, so IRB or equivalent approval is not applicable.

\item {\bf Declaration of LLM usage}
    \item[] Question: Does the paper describe the usage of LLMs if it is an important, original, or non-standard component of the core methods in this research? Note that if the LLM is used only for writing, editing, or formatting purposes and does \emph{not} impact the core methodology, scientific rigor, or originality of the research, declaration is not required.
    \item[] Answer: \answerYes{}.
    \item[] Justification: The paper explicitly describes the use of LLMs in the core method and data construction: Qwen3-32B and Nemotron-3-Nano-Omni-30B-A3B-Reasoning-BF16 are the base models; GPT-5 is used as part of the ReasonIF editor and to generate full-trajectory demonstrations for SFT-GPT; GPT-5.4 generates the IFEval training prompts, edits the IFEval response-construction trajectories, and judges IMO-AnswerBench responses (Table~\ref{tab:overview}). Appendices~\ref{app:editor-details}, \ref{app:sft-data}, and \ref{app:ifeval-transfer} describe these uses.

\end{enumerate}

\end{document}